\documentclass[times, review, 10pt]{elsarticle}

\usepackage{amssymb}
\usepackage{amsmath}

\usepackage{multicol}
\usepackage{multirow}
\usepackage{booktabs}

\journal{Pattern Recognition}

\begin{document}

\begin{frontmatter}



\title{SAIP-Net: Enhancing Remote Sensing Image Segmentation via Spectral Adaptive Information Propagation}


\author{Zhongtao Wang} 
\ead{wangzhongtao@stu.pku.edu.cn}

\author{Xizhe Cao} 

\author{Yisong Chen\corref{cor1}} 
\cortext[cor1]{Corresponding author.}
\ead{chenyisong@pku.edu.cn}

\author{Guoping Wang} 

\affiliation{organization={School of Computer Science, Peking University},
            addressline={No.5 Yiheyuan Road, Haidian District}, 
            city={Beijing},
            postcode={100871}, 
            country={P.R. China}}

\begin{abstract}
 Semantic segmentation of remote sensing imagery demands precise spatial boundaries and robust intra-class consistency, challenging conventional hierarchical models. To address limitations arising from spatial domain feature fusion and insufficient receptive fields, this paper introduces SAIP-Net, a novel frequency-aware segmentation framework that leverages Spectral Adaptive Information Propagation. SAIP-Net employs adaptive frequency filtering and multi-scale receptive field enhancement to effectively suppress intra-class feature inconsistencies and sharpen boundary lines. Comprehensive experiments demonstrate significant performance improvements over state-of-the-art methods, highlighting the effectiveness of spectral-adaptive strategies combined with expanded receptive fields for remote sensing image segmentation. Our code is available at https://github.com/ZhongtaoWang/SAIP-Net.
\end{abstract}


\end{frontmatter}

\begin{figure}
  \centering
  \includegraphics[width=\textwidth]{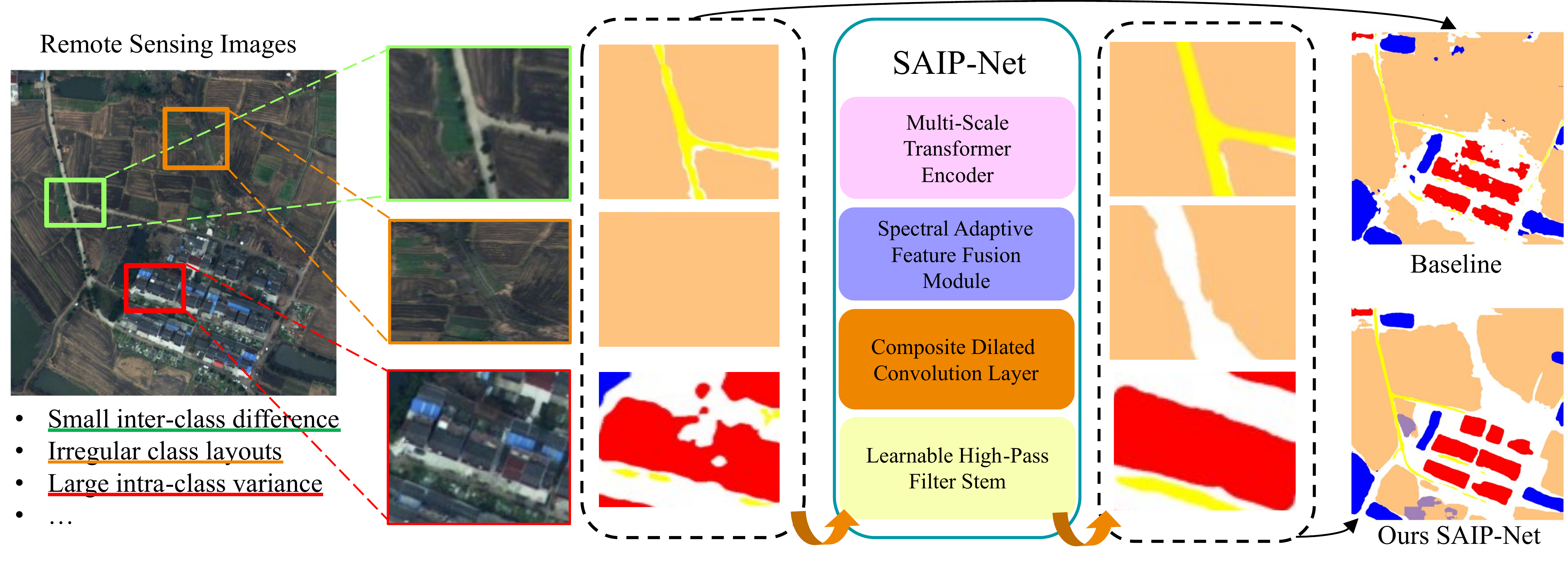}
  \caption{Overview of the challenges and motivations behind SAIP-Net. Remote sensing images often exhibit large intra-class variance, small inter-class differences, and irregular class layouts. To address these challenges, SAIP-Net uses the combination of four modules to improve intra-class consistency and enhance boundary accuracy, significantly improving the segmentation of complex structures in remote sensing images.}
  \label{fig:teaser}
\end{figure}



\section{Introduction}

Semantic segmentation of remote sensing images is a fundamental yet challenging task with profound implications in numerous critical applications, such as land-use classification, urban planning, agriculture monitoring, environmental assessment, and disaster management. Despite significant advances achieved through deep learning techniques, remote sensing imagery continues to pose distinct challenges that conventional hierarchical deep neural networks struggle to address adequately. Such challenges include diverse and complex textural features, significant variations in object scales, intricate boundaries, and the coexistence of highly structured urban areas alongside irregular natural landscapes.

Existing segmentation approaches typically rely on hierarchical feature fusion methods operating primarily within the spatial domain. These conventional fusion methods often directly combine multi-scale feature representations obtained from various hierarchical levels through simple addition or concatenation. However, this spatial-domain fusion frequently leads to significant intra-class inconsistencies, primarily due to disturbances caused by unregulated high-frequency information. Moreover, such methods are typically hindered by boundary ambiguities and inaccuracies arising from blurred details and insufficient contextual information captured by limited receptive fields.

To overcome these limitations, this paper introduces SAIP-Net, an innovative frequency-aware segmentation framework explicitly designed for remote sensing image analysis. At the core of our method is the novel concept of Spectral Adaptive Information Propagation, a mechanism that dynamically manipulates frequency-domain characteristics of image features. As shown in Figure~\ref{fig:teaser}, SAIP-Net systematically employs adaptive spectral filtering techniques to mitigate disruptive high-frequency noise within semantic object regions, thereby significantly enhancing intra-class consistency. Simultaneously, our method incorporates frequency-aware processes to recover and sharpen high-frequency boundary details typically diminished during conventional downsampling and upsampling operations. Beyond frequency-domain enhancements, SAIP-Net also introduces strategies to substantially enlarge the receptive fields of feature extraction modules. Enlarged receptive fields facilitate more comprehensive contextual understanding by capturing extensive spatial interactions and dependencies, crucial for correctly identifying semantic classes at varying scales. By integrating frequency-domain processing and spatial contextual awareness, SAIP-Net significantly improves both feature representation consistency and boundary delineation accuracy.

In conclusion, our contributions can be summarized as:
\begin{itemize}
    \item We introduced SAIP-Net, a frequency-aware segmentation framework that incorporates the novel Spectral Adaptive Information Propagation mechanism to address the inherent challenges of remote sensing image analysis. 
    \item By adaptively propagating spectral information and enlarging the receptive fields, our approach effectively eliminates disruptive high-frequency information within intra-class regions while preserving essential frequency-domain features along class boundaries, which results in substantially improved intra-class consistency and enhanced boundary accuracy.
    \item Experimental results demonstrate that our proposed method significantly outperforms the baseline in semantic segmentation task on remote sensing images.
\end{itemize}

\section{Related Work}
\label{sec:RelatedWorks}

Semantic segmentation assigns semantic labels to each pixel and is widely applied in both natural and remote sensing images. While the two domains share foundational techniques, their data characteristics and task constraints differ significantly. Natural image segmentation focuses on perspective-view scenes with diverse objects and consistent contexts~\cite{long2015fully}, while remote sensing imagery features large-scale, top-down views with high intra-class variability, inter-class similarity, which requires domain-specific solutions~\cite{zhu2017deep}. 

\subsection{Semantic Segmentation in Natural Images}

Early methods, such as thresholding and region growing, were limited by image quality and computational constraints. The advent of deep learning, particularly Convolutional Neural Networks (CNNs) \cite{lecun1998gradient} and Fully Convolutional Networks (FCNs) \cite{long2015fully}, enabled end-to-end pixel-level prediction, forming the basis for encoder-decoder architectures like U-Net \cite{ronneberger2015u} and SegNet \cite{badrinarayanan2017segnet}. These models extract high-level semantic features while preserving spatial resolution.

Subsequent research addressed challenges such as class imbalance, small object detection, and complex backgrounds. Dilated convolutions \cite{li2018csrnet} expanded receptive fields, while PSPNet \cite{zhou2019fusion} introduced pyramid pooling to aggregate multi-scale contextual information. Attention mechanisms, such as Coordinate Attention \cite{hou2021coordinate}, further refine segmentation, especially for small or complex objects. Additionally, unsupervised and semi-supervised approaches \cite{ahn2019weakly,lin2024semples} have been explored to reduce reliance on large annotated datasets.

\subsection{Semantic Segmentation in Remote Sensing Images}

Remote sensing segmentation faces unique challenges, including large-scale scenes, spectral diversity, and subtle intra-class variation. While FCN-based methods \cite{chen2018encoder,sun2019deep,li2021pointflow,xue2022aanet,factseg2022,hou2022bsnet} remain prevalent, additional modules have been developed to enhance spatial and contextual feature extraction. Techniques like channel attention \cite{hu2018squeeze} and region-aware methods, such as S-RA-FCN \cite{mou2020relation} and HMANet \cite{niu2021hybrid}, address feature redundancy and spatial dependencies. Edge-aware strategies, like Edge-FCN \cite{he2020remote}, improve boundary precision by incorporating edge detection.

Recently, Transformer-based models \cite{wang2022novel,xu2023rssformer,xie2021segformer,wang2022unetformer,chen2021transunet,sun2022ringmo} have gained attention for their ability to model long-range dependencies via self-attention. For instance, RSSFormer \cite{xu2023rssformer} introduces adaptive feature fusion to reduce background noise and enhance object saliency. However, Transformers still face challenges in capturing fine local details and accurately delineating object boundaries.

\subsection{Frequency-domain Methods for Semantic Segmentation}

Frequency-domain feature processing has emerged as a promising direction in semantic segmentation. Shan et al. decomposed images via the Fourier transform into high and low frequency components \cite{shan2021decouple}, which capture edge and body information respectively, ensuring object consistency and edge supervision through deep fusion. Zhang et al. proposed FsaNet \cite{zhang2023fsanet}, leveraging a frequency self-attention mechanism on low-frequency components to drastically reduce computational costs. Chen et al. introduced Frequency-Adaptive Dilated Convolution (FADC) \cite{chen2024frequency1}, dynamically adjusting dilation rates by local frequency patterns and enhancing feature bandwidth/receptive fields via adaptive kernels and frequency selection. Their FreqFusion \cite{2024freqfusion} further improves boundary sharpness and feature coherence for dense prediction tasks. Ma et al. developed AFANet \cite{ma2025afanet}, optimizing semantic structures with a CLIP-guided spatial adapter to boost weakly-supervised few-shot segmentation. These advances underscore the transformative potential of frequency-aware strategies in semantic segmentation.

Building on these works, our work proposes a novel approach named SAIP-Net. By adaptively propagating spectral information and enlarging the receptive fields, our approach substantially improved intra-class consistency and enhanced boundary accuracy, thereby significantly improving the segmentation of complex structures in remote sensing images.

\section{Methods}

\begin{figure*}[!t]
    \centering
    \includegraphics[width=\linewidth]{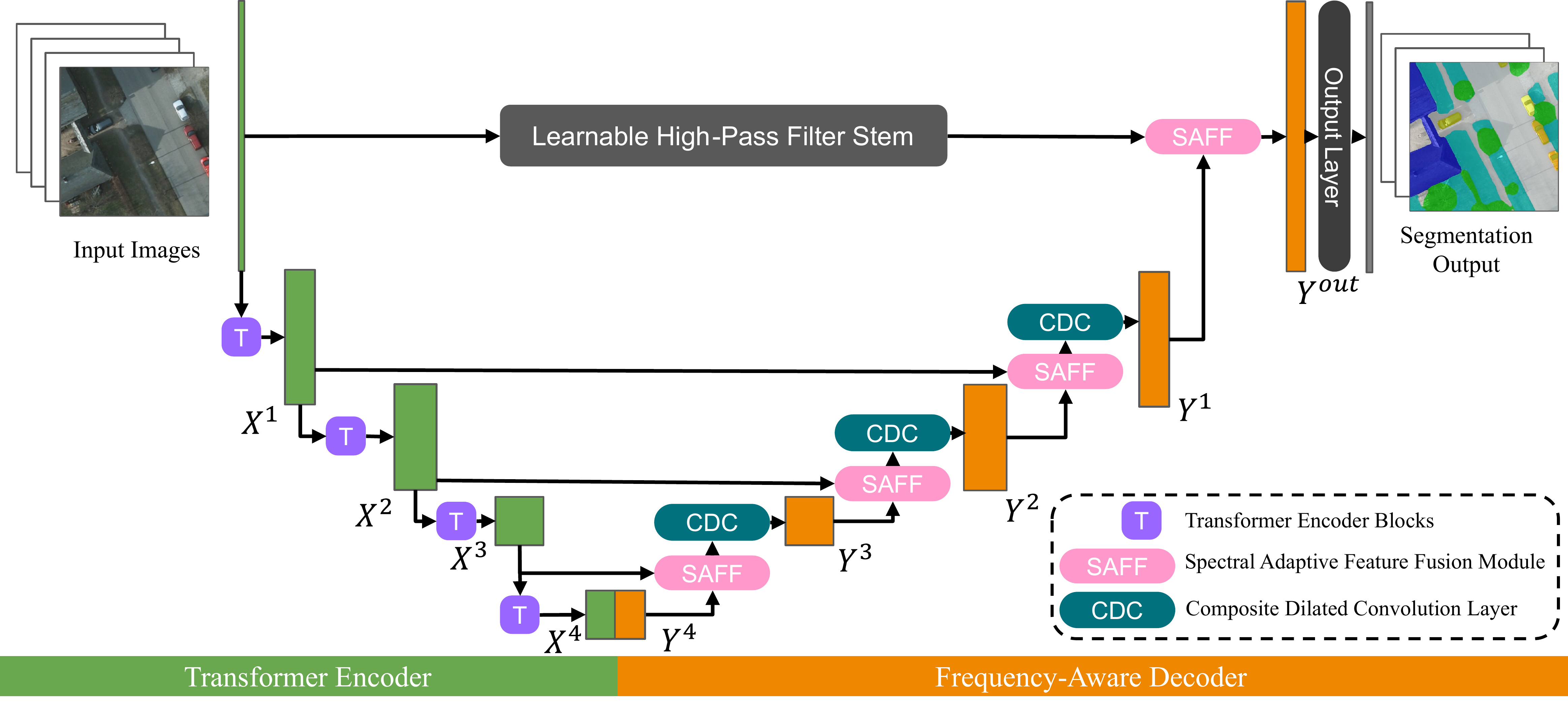}
    \caption{Overview of the proposed SAIP-Net architecture. The network combines: a Learnable High-Pass Filter Stem to enhance edge details, a Transformer Encoder that extracts global context via multi-stage learnable pooling, and a Frequency-Aware Decoder that fuses high- and low-level features using SAFF modules, while integrated Composite Dilated Convolution Layers expand the receptive field. 
    These modules ultimately lead to result in improved intra-class consistency and enhanced boundary accuracy, thereby significantly improving the segmentation of complex structures in remote sensing images.
    }
    \label{fig:main}
\end{figure*}

\label{sec:methods_overview}

Figure~\ref{fig:main} illustrates the overall architecture of our proposed SAIP-Net, which integrates several novel components to address the challenges of remote sensing image segmentation. Our method combines a transformer encoder, a frequency-aware decoder equipped with Spectral Adaptive Feature Fusion (SAFF) module and Composite Dilated Convolution (CDC) Layers, and a Learnable High-Pass Filter Stem (LhpfStem) into a unified framework. The encoder efficiently captures global context through multi-stage feature extraction and learnable pooling. Then the decoder fuses high- and low-level features via the Spectral Adaptive Feature Fusion module to enhance intra-class consistency and boundary precision. The CDC layers in decoder leverage parallel dilated convolutions with varying receptive fields to extract multi-scale spatial details. The LhpfStem suppresses background interference by emphasizing high-frequency edge information at the very beginning of the network. Together, these components enable SAIP-Net to effectively integrate spatial and frequency domain information, resulting in improved segmentation performance under the complex conditions typical of remote sensing imagery.

\subsection{Transformer Encoder}
\label{sec:mvit_encoder}

Our transformer encoder is designed to balance global context modeling and computational efficiency via pooling attention and hierarchical feature extraction. The encoder consists of $4$ stages. Following \cite{li2021improved}, each stage of our encoder includes multiple Transformer blocks, and spatial downsampling between stages is achieved through learnable pooling operations.

\paragraph{Patch Embedding}
The encoder starts by applying a strided convolution (patch embedding layer) on the input image of size $H \times W \times 3$, generating token embeddings of shape $\frac{H}{p} \times \frac{W}{p} \times C$. This effectively partitions the image into non-overlapping patches and projects each to a $C$-dimensional embedding. A smaller patch size $p$ preserves local details crucial for dense prediction.

\paragraph{Transformer Encoder Blocks}
Each block follows the standard structure with multi-head self-attention and MLP, each preceded by LayerNorm and followed by residual connections. We use pooling attention to reduce spatial resolution within attention computation. Specifically, input $X_{in} \in \mathbb{R}^{L \times D}$ is projected and pooled as:
\begin{equation}
Q = \mathcal{P}_Q(X_{in}W_Q), \quad K = \mathcal{P}_K(X_{in}W_K), \quad V = \mathcal{P}_V(X_{in}W_V),
\end{equation}
where $\mathcal{P}\cdot$ denotes a learnable pooling operator, reducing the length from $L$ to $\Tilde{L}$.

We also incorporate relative positional embeddings to maintain spatial awareness and employ a residual connection on the query branch. The output at the l-th stage $X_l$ can be calculated as:
\begin{equation}
X_l = \mathrm{Attn}(Q, K, V) + Q = \mathrm{Softmax}\left(\frac{QK^\top + E^{(\mathrm{rel})}}{\sqrt{D}}\right)V + Q
\end{equation}

where $E^{(\mathrm{rel})}{ij} = Q_i \cdot R{p(i),p(j)}$.
Here, $Q_i$ represents the query vector at token position $i$, and $R_{p(i),p(j)}$ denotes the learnable relative positional embedding between tokens $i$ and $j$, where $p(i)$ and $p(j)$ map token indices to their spatial locations. The inner product $Q_i \cdot R_{p(i),p(j)}$ injects relative position bias into the attention logits. To reduce the number of learnable parameters, we follow~\cite{li2021improved} and decompose $R_{p(i),p(j)}$ along spatial dimensions as:
\begin{equation}
    R_{p(i), p(j)} = R^{\mathrm{h}}_{h(i), h(j)} + R^{\mathrm{w}}_{w(i), w(j)},
\end{equation}
where $R^{\mathrm{h}}$ and $R^{\mathrm{w}}$ are independent learnable embeddings along the height and width axes, and $h(i)$, $w(i)$ refer to the row and column indices of token $i$. This decomposition reduces the embedding complexity from $\mathcal{O}(HW)$ to $\mathcal{O}(H + W)$.

\paragraph{Hierarchical Structure}
Like most multi-scale image encoders, our encoder employs hierarchical pooling across stages. The spatial size is reduced, and the channel and dimension at each stage are increased at $\frac{H}{4} \times \frac{W}{4} \times C$, $\frac{H}{8} \times \frac{W}{8} \times 2C$, $\frac{H}{16} \times \frac{W}{16} \times 4C$
and $\frac{H}{32} \times \frac{W}{32} \times 8C$.

This design enables the encoder to capture multi-scale semantics with reduced complexity, making it well-suited for high-resolution remote sensing image segmentation.

\subsection{Spectral Adaptive Feature Fusion Module}
\label{sec:SAFF}

\begin{figure}[t]
    \centering
    \includegraphics[width=\linewidth]{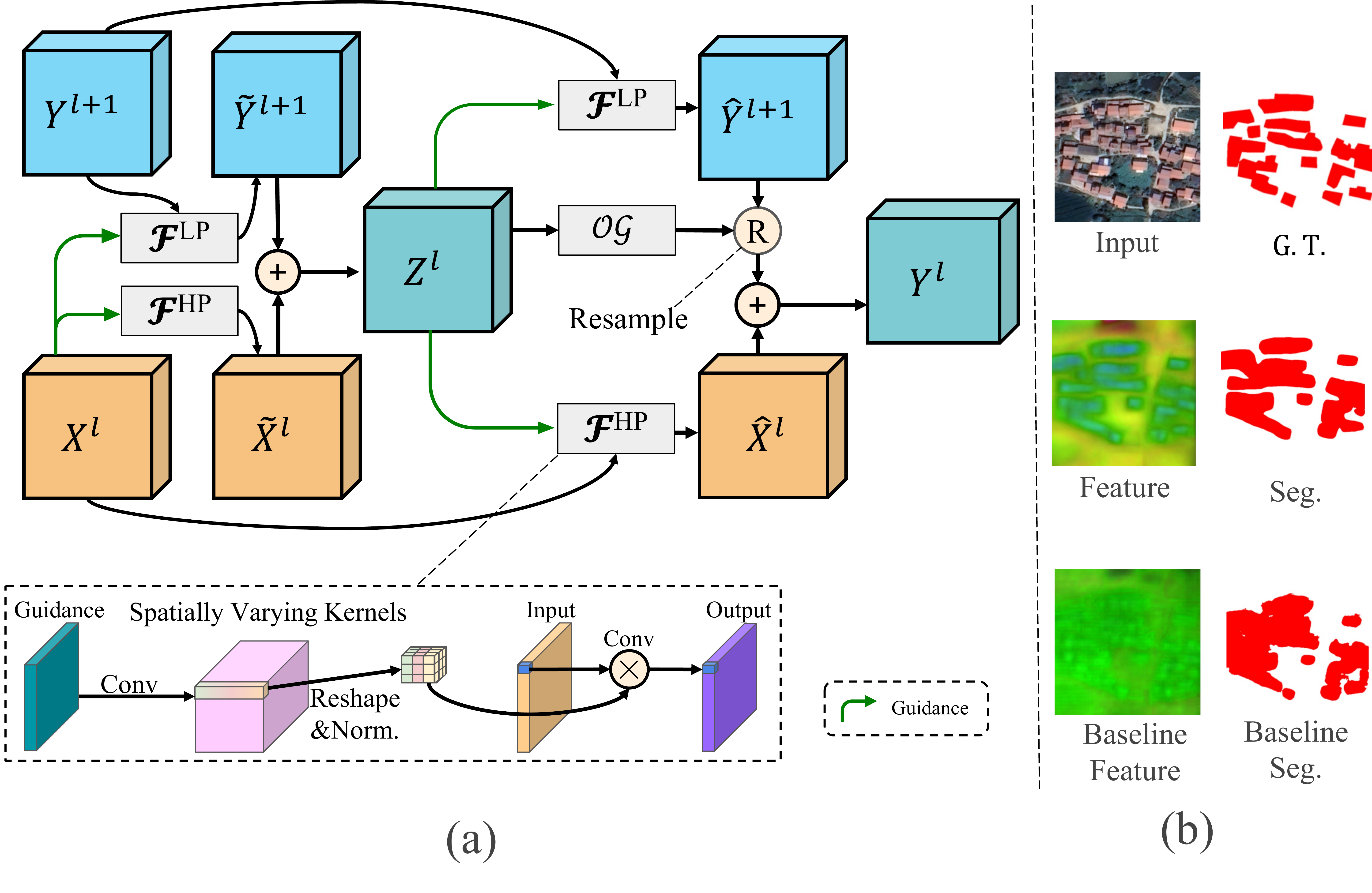}
    \caption{(a). Illustration of Spectral Adaptive Feature Fusion (SAFF) module. The module integrates high-level and low-level features using content guided low-pass and high-pass filters alongside spatial offset estimation. The structure of $\mathcal{F}^{\text{HP}}$ is enlarged at the bottom while $\mathcal{F}^{\text{LP}}$ shares a similar architecture. (b). Our module enhances intra-class consistency and boundary accuracy at the feature level, leading to better segmentation results.}
    \label{fig:ff}
\end{figure}

To fully utilize the multi-scale features extracted by the encoder, we propose a Spectral Adaptive Feature Fusion Module (SAFF Module). This module is guided by high- and low-level features respectively, leverages frequency-domain priors to improve feature alignment and semantic consistency during upsampling, and address the challenges of intra-class inconsistency and boundary degradation. 

Inspired by \cite{2024freqfusion,he2012guided,ma2019carafe}, incorporating a guidance signal (e.g., low-level features) provides valuable local structural priors. These priors help direct the filtering process such that the spatially varying kernels can adapt based on local content, ensuring that essential details like edges and textures are accurately preserved while undesired noise is suppressed. 
Additionally, guided filters that vary at each position allow fine-grained feature reassembly and precise spatial alignment, ensuring that the fused features accurately reflect both the global context and the local details. Such flexibility directly contributes to improved overall segmentation quality and enhanced boundary delineation.

As illustrated in Figure~\ref{fig:ff} (a), our SAFF module begins with an initial fusion step that integrates complementary high-level and low-level features. First, the low-level feature $\mathbf{X}^{l}$ is enhanced by a content guided high-pass filter $\mathcal{F}^{\text{HP}}$ guided by $\mathbf{X}^{l}$ itself that emphasizes edge details, with a residual connection preserving the original information. In parallel, the high-level feature $\mathbf{Y}^{l+1}$ is processed through a content guided low-pass filter $\mathcal{F}^{\text{LP}}$ guided by $\mathbf{X}^{l}$ to suppress noisy high-frequency components and upsampled to match the resolution of the low-level feature. Then these processed features are fused via element-wise addition to produce the initial fused feature map ${Z}^{l}$:
\begin{equation}
\begin{aligned}
\tilde{\mathbf{Y}}^{l+1} = \mathcal{F}^{\text{LP}}(X^l,\mathbf{Y}^{l+1}), \quad
\tilde{\mathbf{X}}^{l} = \mathcal{F}^{\text{HP}}(X^l,\mathbf{X}^{l}) + \mathbf{X}^{l}, \quad
\mathbf{Z}^{l} = \tilde{\mathbf{Y}}^{l+1} + \tilde{\mathbf{X}}^{l}.
\end{aligned}
\end{equation}

After this, a second fusion is performed by refining the alignment between high-level and low-level features. In particular, we rely on $\mathbf{Z}^{l}$ to perform an additional content guided low-pass filtering on $\mathbf{Y}^{l+1}$ and content guided high-pass filtering on $\mathbf{X}^{l}$:

\begin{equation}
\begin{aligned}
\hat{\mathbf{X}}^{l} = \mathcal{F}^{\text{HP}}(Z^l,\mathbf{X}^{l}) + \mathbf{X}^{l}, \quad
\hat{\mathbf{Y}}^{l+1} = \mathcal{F}^{\text{LP}}(Z^l,\mathbf{Y}^{l+1}).
\end{aligned}
\end{equation}

Specifically, an offset generator predicts spatial offsets $\bigl(u(i,j),v(i,j)\bigr)$ based on local feature similarities. These offsets are used to resample the upsampled, low-pass filtered high-level feature \(\hat{\mathbf{Y}}^{l+1}\), aligning it with the high-pass enhanced low-level feature \(\hat{\mathbf{X}}^{l}\). The secondly fused feature is then obtained by element-wise addition:
\begin{equation}
\label{eq:resamp}
\begin{aligned}
u(i,j) , v(i,j) = \mathcal{OG}(Z^l)_{ij},  \quad \mathbf{Y}^{l}_{i,j} = \hat{\mathbf{Y}}^{l+1}_{i+u(i,j),j+v(i,j)} + \hat{\mathbf{X}}^{l}_{i,j}.
\end{aligned}
\end{equation}
As shown in Figure~\ref{fig:ff} (b), this process ensures that the feature map benefits from both the smooth global context and the detailed local structure, leading to improved segmentation quality.
It also effectively combines the smooth, context-rich high-level features with the detail-preserving low-level features, laying a solid foundation for improved segmentation performance.

The key components of this module are two adaptive filter: Content Guided Low-Pass/High-Pass Filter $\mathcal{F}^{\text{LP}}$/$\mathcal{F}^{\text{HP}}$ and an offset generator $\mathcal{OG}$. We elaborate on each component as follows:

\paragraph{Content Guided Low-Pass/High-Pass Filter}
Inspired by \cite{he2012guided,ma2019carafe}, we take $\mathbf{Z}$ as the guidance feature to predict the spatially variant filter weights filtering feature map $\mathbf{X}$. The filter consists of several steps:

\begin{enumerate}
    \item \textbf{Filter Prediction.}  
    A $3\times3$ convolution predicts spatially variant filters $\mathbf{\bar V}^{l}$ from $\mathbf{Z}$ and normalizes them using softmax. The softmax operation imposes a low-pass behavior on the convolutional kernel by concentrating weights on nearby positions and suppressing distant responses, thus favoring smooth, low-frequency features. For the high-pass filter, it is then subtracted from an identity kernel $\mathbf{E}$ to obtain high-pass behavior:
    \begin{equation}
    \resizebox{0.9\linewidth}{!}{$
        \mathbf{\bar V} 
        = \text{Conv}_{3\times3}\!\bigl(\mathbf{Z}\bigr), 
        \quad
        \mathbf{\bar W}^{p,q}_{i,j} 
        = 
        \begin{cases}
        \frac{\exp(\mathbf{\bar V}^{p,q}_{i,j})}{\sum_{p,q\in \Omega}\exp(\mathbf{\bar V}^{p,q}_{i,j})}, & \text{(low-pass)}\\
         \mathbf{E}\;-\; \frac{\exp(\mathbf{\hat V}_{i,j}^{p,q})}{\sum_{p,q\in\Omega}\exp(\mathbf{\hat V}_{i,j}^{p,q})}, & \text{(high-pass)} \\
        \end{cases}
    $}
    \end{equation}
    where $\Omega$ is the $K\times K$ convolution window (e.g., $K=3$).

    \item \textbf{Spatially Variant Convolution.}  
    The high-level feature $\mathbf{Y}^{l+1}$ is convolved with the predicted weights:
    \begin{equation}
        \mathbf{Y}^{g}_{i,j} 
        = \sum_{p,q\in \Omega}\mathbf{\bar W}^{g,p,q}_{i,j}\,\mathbf{X}_{i+p,\,j+q},
    \end{equation}
    where channels are grouped ($g=1,\dots,4$) following sub-pixel convolution patterns.

    \item \textbf{Upsampling of High-Level Features.}  
    If the input is high-level features, the four groups $\mathbf{Y^g}$ are rearranged using Pixel Shuffle\cite{2016pixelshuffle} to obtain a smooth $2\times$ upsampled feature map $\mathbf{Y} \in \mathbb{R}^{C\times 2H \times 2W}$.
    
\end{enumerate}

\paragraph{Offset Generator.}
For further refinement, especially near complex boundaries, the offset generator  $\mathcal{OG}(Z)$ estimates spatial displacements for further resampling step under the guidance of input feature map $Z$ :

\begin{enumerate}
    \item \textbf{Local Similarity Computation.}  
    Cosine similarity is computed between a pixel and its neighbors:
    \begin{equation}
        \mathbf{S}^{p,q}_{i,j} 
        = \frac{\sum_{c} \mathbf{Z}_{c,i,j}\,\mathbf{Z}_{c,i+p,j+q}}
        {\sqrt{\sum_{c} (\mathbf{Z}_{c,i,j})^2}\;\sqrt{\sum_{c} (\mathbf{Z}_{c,i+p,j+q})^2}},
    \end{equation}

    \item \textbf{Offset Prediction.}  
    Concatenate $\mathbf{Z}$ and $\mathbf{S}$, and apply two convolutions to obtain directional offset $\mathbf{D}$ and magnitude $\mathbf{A}$:
    \begin{equation}
        \mathbf{D} 
        = \text{Conv}_{3\times3}([\mathbf{Z}, \mathbf{S}]),
        \quad
        \mathbf{A} 
        = \sigma\!\bigl(\text{Conv}_{3\times3}([\mathbf{Z}, \mathbf{S}])\bigr),
    \end{equation}
    Where $[...]$ denotes the concat operation and $\sigma$ denotes the sigmoid function.
    Finally, the offset is given by $\mathbf{O} = \mathbf{D} \cdot \mathbf{A}$.

    \item \textbf{Spatial Resampling.}  
    As in Equation~\ref{eq:resamp}, high-level features $\mathbf{\tilde{Y}}$ are resampled using the offsets $\mathbf{O}$ to correct class-wise inconsistencies and refine boundaries.
\end{enumerate}






\subsection{Composite Dilated Convolution Layer}
\label{sec:CDC}

\begin{figure}[t]
    \centering
    \includegraphics[width=\linewidth]{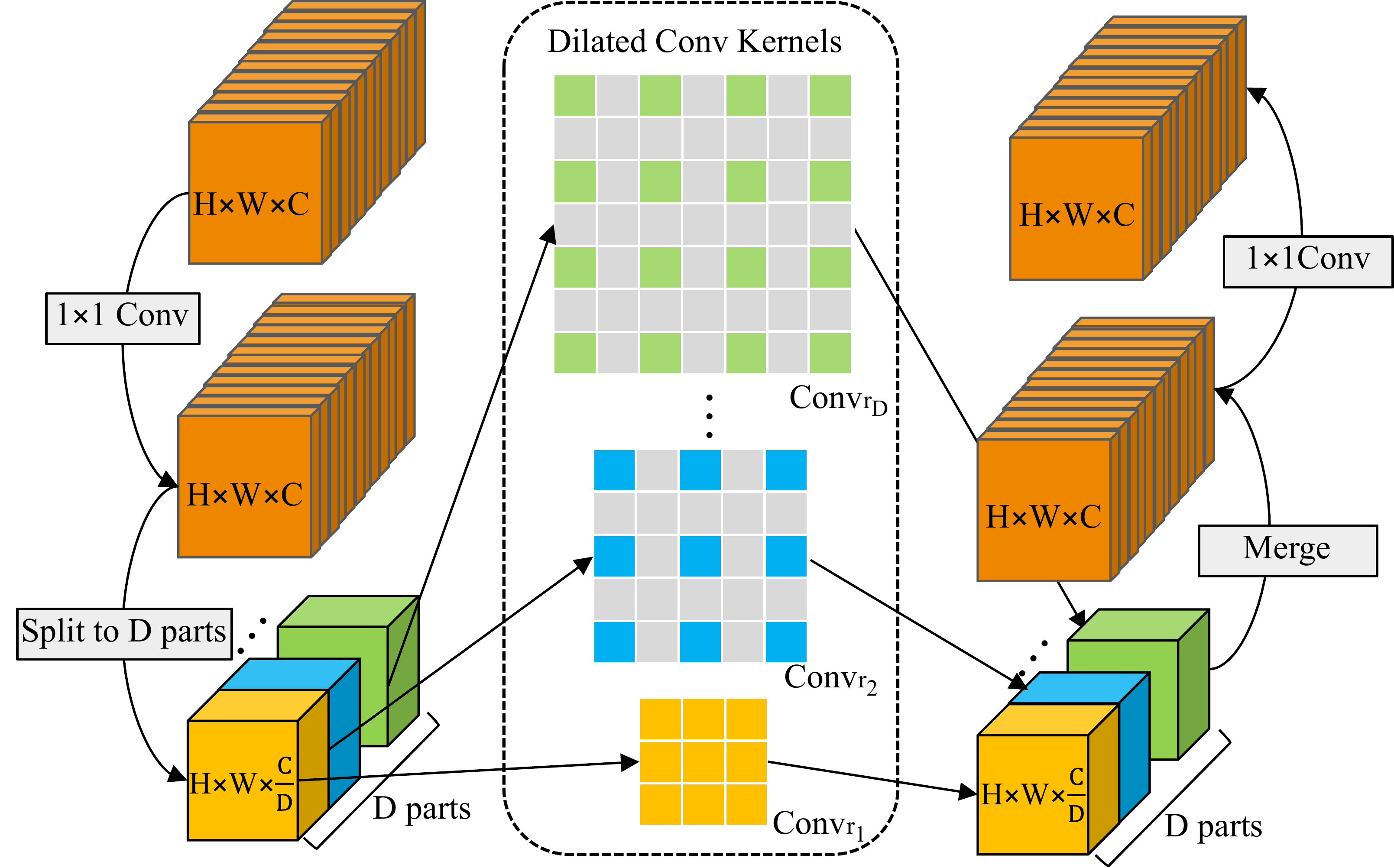}
    \caption{Composite Dilated Convolution Framework: After a channel-mixing operation, we split the feature map into multiple parts processed with different dilation rates, and the resulting multi-scale features are concatenated, refined, and upsampled to reconstruct high-resolution segmentation maps.}
    \label{fig:CDConv}

\end{figure}

To expand the receptive field at low cost, we design a Composite  Dilated Convolution Layer that processes features in a multi-scale manner. As shown in Figure~\ref{fig:CDConv}, this framework is built upon a cascade of operations that first integrates channel information, then diversifies spatial context via parallel dilated convolutions, and finally refines the combined feature representation.

Given an input feature map $Y^l$, which fuses high-level semantic information from deeper layers, we begin with a lightweight channel integration. This is achieved using a $1\times 1$ convolution that acts as a channel-mixer, blending the channel information without altering the spatial dimensions. This preliminary mixing primes the feature map for subsequent multi-scale processing.

The integrated feature $\mathbf{x}$ is then partitioned equally into $D$ sub-tensors, denoted as $\mathbf{x}_1 \ldots \mathbf{x}_D$. Each sub-tensor is independently processed by a dilated convolution operation with a specific dilation rate $d_i$.
The receptive field size of each branch is calculated as $r_i = d_i (k - 1) + 1$, where larger $k$ for larger receptive fields and smaller $k$ for smaller ones.
The individual outputs are given by:
\begin{equation}
\mathbf{y}_i = \operatorname{Conv}_{r_i}(\mathbf{x}_i), \quad i \in \{1 \ldots D\}. 
\end{equation}
These outputs are then concatenated to form a multi-scale feature representation. Then the concatenated features are further refined through a merging module. This module employs a series of convolutions—typically a combination of $1\times 1$ and $3\times 3$ kernels—each followed by Batch Normalization and ReLU activation. This post-mixing step not only fuses the diverse scale information but also introduces non-linearity to enhance the discriminative capability of the feature representation. The entire Composite Dilated Convolution operation can be summarized as:
\begin{equation}
\mathcal{F}_{\text{CDC}}(Y^l) = \operatorname{Conv}\Big( \Big[ y_1 \ldots y_D \Big] \Big).
\end{equation}

To restore the original spatial resolution, the output of the Composite Dilated Convolution is passed to an upsampling block. 
Through extensive tuning and experiments, we found that setting $D=3$ provides an optimal balance between effectively capturing diverse contextual information and maintaining a reasonable model complexity.


\subsection{Learnable High-Pass Filter Stem}
\label{sec:stem}


Remote sensing images feature intricate high-frequency details (e.g., boundaries) alongside low-frequency noise like clouds and shadows, requiring frequency-domain guidance during feature extraction. The SAFF module (Section~\ref{sec:SAFF}) depends on well-formed feature maps, but the raw 3-channel RGB input at the network's top resides in a shallow, low-dimensional space which lacks the rich, abstract features present in deeper latent representations. So we need to extract useful information for reliably generating dynamic convolution kernels.

To address this, we propose the \textbf{Learnable High-Pass Filter Stem (LhpfStem)}, a dedicated module positioned at the residual connection of the network’s top layer.  It employs fixed, learnable convolution kernels that are optimized during training to perform high-pass filtering directly on the raw input at low cost.



\begin{figure}[t]
    \centering
    \includegraphics[width=0.99\linewidth]{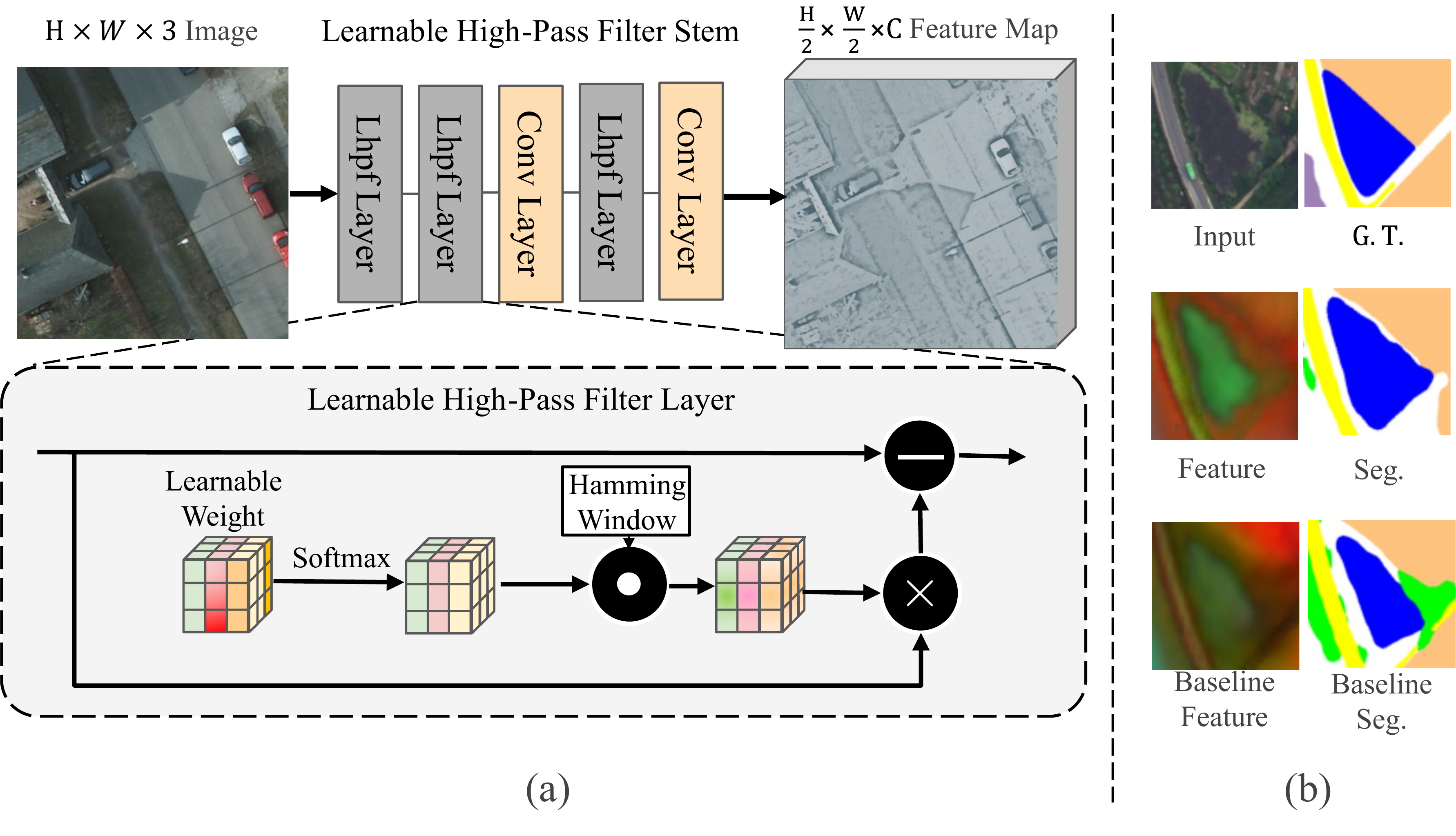}
    \caption{(a). Overview of the proposed \textbf{LhpfStem}. The module is composed of a stack of Lhpf layers. In a Lhpf layer, the high-pass output is computed by subtracting the low-pass response features from the input, thereby enhancing edge details and fine structures. (b). Our module extracts key high-frequency features, sharpens feature boundaries, and improves segmentation performance.}
    \label{fig:stem}
\end{figure}

 As shown in Figure~\ref{fig:stem} (a), our LhpfStem is designed as a stack of Learnable High-Pass Filter Layers (Lhpf Layers) and interspersed with convolutional layers. In each Lhpf layer, given the input image or feature map \( X \in \mathbb{R}^{C \times H \times W} \), the module uses a learnable weight \( W_{\text{l}} \) of shape \((C, 1, K^2)\) to perform adaptive
high-pass filtering. The weights are firstly normalized using a softmax over the kernel dimension. A hamming window \( H \in \mathbb{R}^{K \times K} \) is then applied to modulate these weights, and the result is re-normalized:
\begin{equation}
\tilde{W}(p,q) = \frac{ \operatorname{Softmax}(W_{\text{l}})(p,q) \cdot H(p,q)}{\sum_{p',q'\in \Omega} \bar{W}(p',q') \cdot H(p',q')},
\end{equation}
where \(\Omega\) denotes the set of kernel indices. The modulated kernel \(\tilde{W}\) is reshaped to a convolution kernel of size \((C,1,K,K)\) and used in a depthwise convolution on \( x \) to obtain the low-pass filtered output:
\begin{equation}
\tilde{X}(i,j) = \sum_{p,q \in \Omega} X(i+p,j+q)\, \tilde{W}(p,q).
\end{equation}
Finally, the high-pass filtered output is computed by subtracting the low-pass response from the original input:
\begin{equation}
\text{HighPass}(x)(i,j) = X(i,j) - \tilde{X}(i,j).
\end{equation}

As shown in Figure~\ref{fig:stem} (b), this design not only enhances edge details and preserves fine structures but also provides robust frequency guidance to subsequent layers. Consequently, LhpfStem effectively overcomes the limitations of using raw images for adaptive filtering, ensuring that critical high-frequency information is retained and enhancing the overall segmentation performance in complex remote sensing scenarios.

The LhpfStem offers a significant \textbf{low-cost} advantage compared to transformer-based models or deeper convolutional networks. By relying on fixed, learnable convolution kernels and depthwise convolutions for high-pass filtering, the LhpfStem avoids the extensive computational overhead and large memory footprint typically associated with those alternative architectures.

To generate the final segmentation prediction, we fuse $X^{hp} = \text{LhpfStem}(x)$ with the top-level features from the decoder \(Y^1\) into \(Y_{out}\) via a SAFF module. Finally, an output layer is applied to \(Y_{out}\) to produce the segmentation logits, and a softmax activation converts these logits into pixel-wise class probabilities.

\subsection{Loss Functions}
\label{sec:loss_function}







To train SAIP-Net, we adopt a hybrid loss that combines Cross-Entropy (CE) and Dice\cite{milletari2016v} losses, balancing pixel-wise classification and region-level overlap. The total loss is defined as:

\begin{equation}
    \mathcal{L}_{\text{total}} = \lambda \left(-\frac{1}{N}\sum_{i=1}^{N}\sum_{c=1}^{C} y_{i,c} \log(\hat{y}_{i,c}) \right) + (1 - \lambda) \left(1 - \frac{2\sum_{i,c} y_{i,c} \hat{y}_{i,c}}{\sum_{i,c} y_{i,c} + \sum_{i,c} \hat{y}_{i,c}} \right),
\end{equation}
where \(N\) is the number of pixels, \(C\) the number of classes, \(y_{i,c}\) the ground truth, and \(\hat{y}_{i,c}\) the predicted probability. This formulation promotes both accurate classification and spatial consistency.

\section{Experiments}
\label{sec:ExperimentalResults}

\subsection{Implementation Details}
\label{sec:impl_detail}

\paragraph{Dataset Selection.} To evaluate our model, we conduct experiments on two widely used remote sensing datasets: Potsdam\footnote{https://www.isprs.org/education/benchmarks/UrbanSemLab/2d-sem-label-potsdam.aspx} and LoveDA\cite{wang2021loveda}. All results are reported on the test/validation set of LoveDA and the test set of Potsdam, which are strictly held out during training and model selection to avoid data leakage. 


\paragraph{Settings.} Our model is trained with an initial learning rate of 0.00006, utilizing a warm-up and polynomial decay strategy to adjust the learning rate during training. The loss weight balancing factor is set to $\lambda=0.5$. A batch size of 16 is employed to balance memory consumption and training speed. The Adam optimizer is used with momentum parameters set to 0.9 and 0.999. All experiments are performed on a server equipped with an NVIDIA RTX 4090 GPU. 


\begin{table}[h]

\centering
\caption{Comparison of model efficiency and computational cost. SAIP-Net achieves a favorable balance between performance and complexity, with reduced computational cost, particularly in GFLOPs ($1024\times1024$ input).}
\renewcommand{\arraystretch}{1.2}
\resizebox{!}{4em}{
\begin{tabular}{l|c|c} 
\toprule
Method & Params(MB)$\downarrow$ & GFLOPs$\downarrow$ \\
\midrule
TransUnet & 90.7 & 233.7 \\
DeepLabV3+ & \textbf{12.47} & 216.85 \\
AerialFormer-T & 42.7 & 192.76 \\
\hline
\textbf{Our SAIP-Net} & 37.21 & \textbf{176.8} \\
\bottomrule
\end{tabular}}

\label{tab:param}
\end{table}

\subsection{Evaluation Metrics.}

\label{sec:eval_metrics}
We compare our method against several baseline approaches \textbf{with consistent or larger parameter magnitudes}, including UNet \cite{ronneberger2015u}, SegNet \cite{badrinarayanan2017segnet}, DeepLabV3+ \cite{chen2018encoder}, DANet \cite{fu2019dual}, LANet \cite{ding2020lanet}, FFPNet \cite{xu2020spatial}, TransUNet \cite{chen2021transunet}, UperNet RSP-Swin-T \cite{wang2022empirical}, RSSFormer \cite{xu2023rssformer}, FactSeg \cite{factseg2022}, UNetFormer \cite{wang2022unetformer}, \cite{yamazaki2023aerialformer}, LSKNet\cite{li2025lsknet} , DecoupleNet \cite{lu2024decouplenet} and LOGCAN++ \cite{ma2025logcanpp}. As shown in Table~\ref{tab:param}, our model achieves comparable or even lower parameter counts and GFLOPs ($1024\times1024$ input) compared to baseline methods by adopting a more efficient backbone, a lightweight upsampling strategy, an effective receptive field expansion mechanism, and a compact stem with fewer parameters. 

For clarity and focus, Table~\ref{tab:param} reports comparisons with a subset of representative methods, including those with strong overall performance and diverse architectural designs. Notably, AerialFormer-T has demonstrated state-of-the-art performance across multiple benchmarks, making it a strong reference for evaluating comprehensive effectiveness. Thus, we include AerialFormer and several other key baselines in the parameter and GFLOPs comparison to highlight the efficiency of our design without loss of generality.

We assess model performance using standard metrics: mean Intersection over Union (mIoU), Overall Accuracy (OA), and mean F1 score (mF1). 


\subsection{Test Results}

\begin{figure}[!t]
    \centering
    \resizebox{\linewidth}{!}{
    \includegraphics{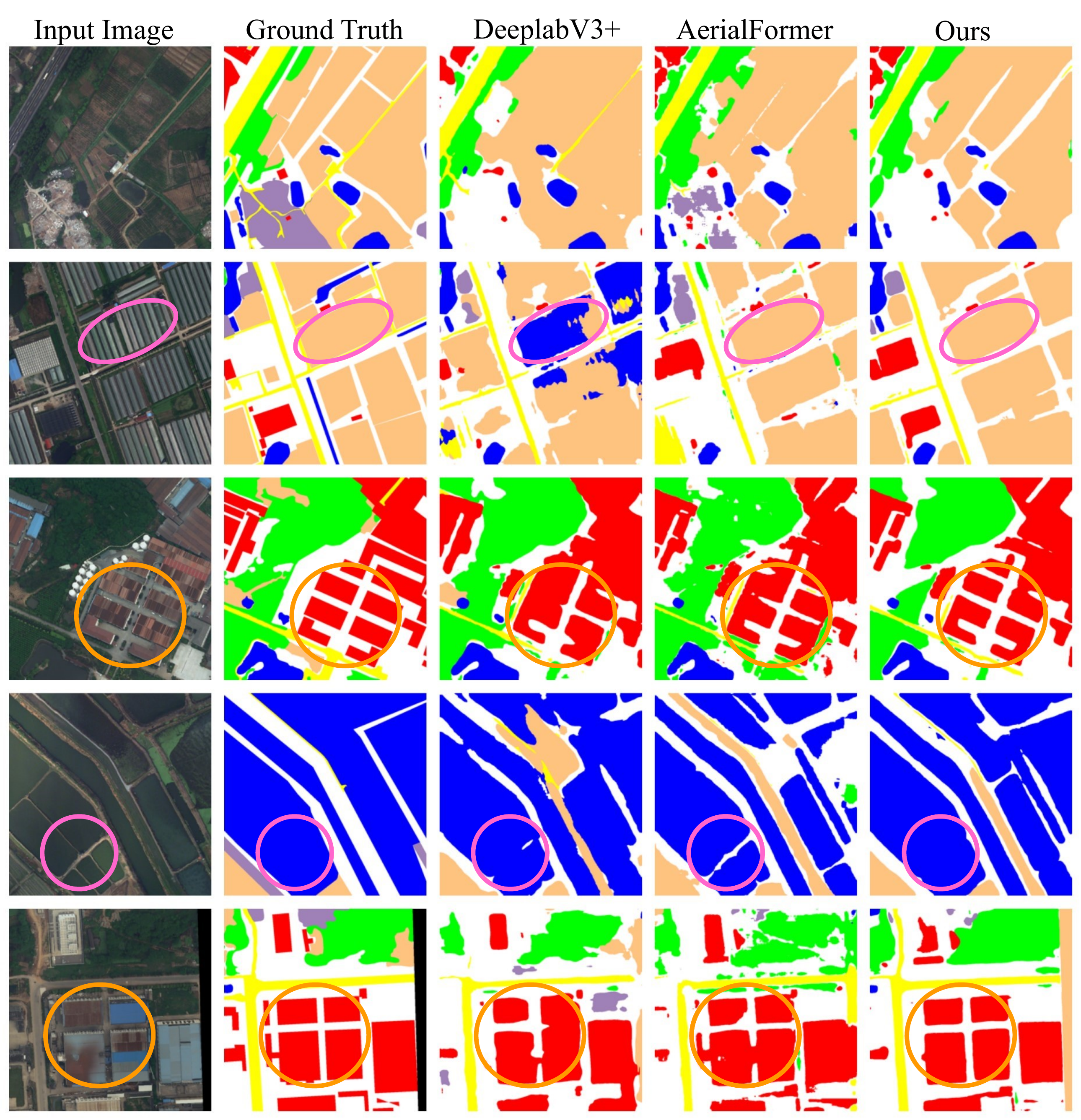}
    }
    \caption{Visual comparison of segmentation results on LoveDA dataset. The regions enclosed by pink ellipses indicate areas with improved intra-class consistency, and the areas marked by orange ellipses highlight enhanced boundary accuracy.}
    \label{fig:result_loveda}
\end{figure}

\begin{figure}[!t]
    \centering
    \resizebox{\linewidth}{!}{
    \includegraphics{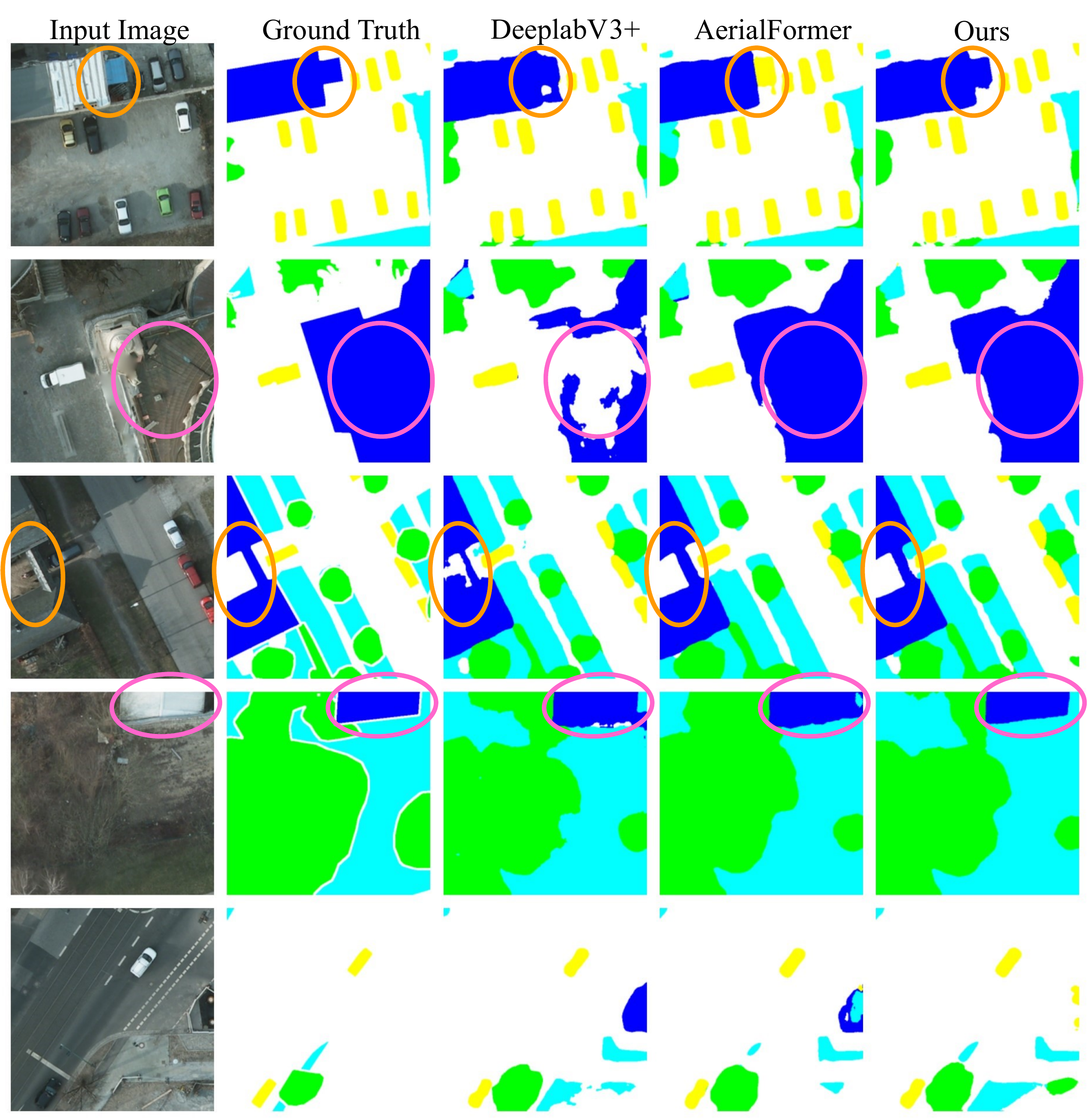}
    }
    \caption{Visual comparison of segmentation results on Potsdam dataset. The regions enclosed by pink ellipses indicate areas with improved intra-class consistency, and the areas marked by orange ellipses highlight enhanced boundary accuracy.}
    \label{fig:result_potsdam}
\end{figure}

Figure~\ref{fig:result_loveda} presents segmentation results on the LoveDA dataset, which includes high-resolution images of both urban and rural scenes with complex semantics. Our method is evaluated on both the validation and test sets of LoveDA, with detailed quantitative results provided in Table~\ref{tab:loveda_val} and Table~\ref{tab:loveda_test}, respectively. The results demonstrate that our model consistently achieves competitive performance across different classes. Visual comparisons further highlight the robustness of our approach in handling diverse object appearances and fine-grained details inherent to remote sensing imagery.

The Potsdam dataset features diverse urban scenes captured under various conditions. As summarized in Table~\ref{tab:potsdam}, our method achieves the highest mIoU, OA, and mF1 scores, demonstrating its capability to handle the complexities of urban remote sensing imagery. Notably, as illustrated in Figure~\ref{fig:result_potsdam}, our approach shows superior performance in capturing fine segmentation details and accurately delineating class boundaries.

\begin{table}[!ht]
\caption{Quantitative results on the official validation set of LoveDA. 
Bold numbers indicate the best performance in each column.}
\centering
\resizebox{\linewidth}{!}{
\begin{tabular}{l|c|ccccccc}
\toprule
\textbf{Method} & \textbf{mIoU $\uparrow$} & \textbf{Bkg.} & \textbf{Building} & \textbf{Road} & \textbf{Water} & \textbf{Barren} & \textbf{Forest} & \textbf{Agri.} \\
\midrule
UNet & 47.33 & 53.95 & 56.13 & 51.26 & 69.15 & 19.75 & 29.00 & 52.10 \\
TransUNet & 50.56 & 53.18 & 62.37 & 52.27 & 69.06 & 27.20 & 34.61 & 55.19 \\
DeepLabV3+ & 50.15 & 51.75 & 59.92 & 53.76 & 68.35 & 26.60 & 39.14 & 51.53 \\
SegNet & 48.34 & 53.02 & 54.53 & 55.42 & 69.08 & 16.37 & 35.67 & 54.31 \\
AerialFormer-T & 52.57 & 53.80 & 64.02 & 55.61 & 66.90 & \textbf{32.83} & \textbf{45.37} & 49.46 \\
\hline 
\textbf{Our SAIP-Net} & \textbf{54.46} & \textbf{55.36} & \textbf{64.53} & \textbf{57.66} & \textbf{71.77} & 28.90 & 43.53 & \textbf{59.44} \\
\bottomrule
\end{tabular}
}
\label{tab:loveda_val}
\end{table}

\begin{table}[!ht]
\caption{Quantitative results on the official test set of LoveDA. Bold numbers indicate the best performance in each column.}
\centering
\resizebox{\linewidth}{!}{
\begin{tabular}{l|c|ccccccc}
\toprule
\textbf{Method} & \textbf{mIoU $\uparrow$} & \textbf{Bkg.} & \textbf{Building} & \textbf{Road} & \textbf{Water} & \textbf{Barren} & \textbf{Forest} & \textbf{Agri.} \\
\midrule
UNet & 47.84 & 43.06 & 52.74 & 52.78 & 73.08 & 10.33 & 43.05 & 59.87 \\
TransUNet & 48.90 & 43.00 & 56.10 & 53.70 & 78.00 & 9.30  & 44.90 & 56.90 \\
DeepLabV3+ & 49.30 & 44.20 & 55.00 & 54.20 & 77.10 & 10.50 & 45.30 & 61.00 \\
SegNet & 46.70 & 41.20 & 51.20 & 50.70 & 72.00 & 9.80  & 42.10 & 58.30 \\
FactSeg  & 48.9 & 42.6 & 53.6 & {{52.8}} & 76.9 & 16.2 & 42.9 & 57.5\\
UNetFormer  & {52.4} & {44.7} & {58.8} &{54.9} & {79.6} & 20.1 & {46.0} & {62.5}\\
RSSFormer-B &  {52.4} & \textbf{52.4} & \textbf{60.7} & {55.21} & {76.29} & {18.73} & {45.39} & {58.33}\\
AerialFormer-T & 52.0 & 45.21 & 57.84 & 56.46 & 79.63 & 19.20 & 46.12 & 59.53 \\
LSKNet-T &  53.2 & 46.4 & 59.5 & 57.1 & 79.9 & 21.8 & 46.6 & 61.4\\
LSKNet-S & \textbf{54.0} & 46.7 & 59.9 & 58.3 & 80.2 & \textbf{24.6} & 46.4 & 61.8 \\
DecoupleNet-D2 & 53.1 & 45.3 & 59.5 & 56.3 & 80.6 & 20.9 & 46.2 & 63.1 \\
LOGCAN++ & 52.0 & 47.37 & 58.38 & 56.46 & 80.05 & 18.44 & 47.91 & \textbf{64.80} \\
\hline 
\textbf{Our SAIP-Net} & 53.30 & 46.32 & 58.64 & \textbf{59.65} & \textbf{81.42} & 15.38 & \textbf{47.33} & 64.33 \\
\bottomrule
\end{tabular}
}
\label{tab:loveda_test}
\end{table}

\begin{table}[!ht]
\caption{Quantitative results on the Potsdam dataset. Bold numbers indicate the best performance in each column.}
\label{tab:potsdam}
\centering
\resizebox{\linewidth}{!}{
\begin{tabular}{l|c|cc|ccccc}
\toprule
\multirow{2}{*}{Method} & \multirow{2}{*}{mIoU $\uparrow$} & \multirow{2}{*}{OA $\uparrow$} & \multirow{2}{*}{mF1 $\uparrow$} & \multicolumn{5}{c}{F1 Score per Class $\uparrow$} \\
\cline{5-9}
 & & & & ImpSurf. & Build. & LowVeg. & Tree & Car \\
\midrule
DeepLabV3+ & 81.69 & 89.60 & 89.79 & 92.27 & 95.52 & 85.71 & 86.04 & 89.42 \\
DANet & 87.28 & 89.72 & 89.14 & 91.61 & 96.44 & 86.11 & 88.04 & 83.54 \\
LANet & 86.20 & 90.84 & 91.95 & 93.05 & 97.19 & 87.30 & 88.04 & 94.19 \\
FFPNet & 86.50 & 91.10 & 92.44 & 93.61 & 96.70 & 87.31 & 88.11 & 96.46 \\
UNetFormer & {86.80} & {91.31}& {92.89} &  {93.62} & {97.24} & {87.73} & {88.91} & {96.52} \\
UperNet RSP-Swin-T & 73.50 & 90.78 & 90.03 & 92.65 & 96.35 & 86.02 & 85.39 & 89.75 \\
AerialFormer-T & 88.51 & 93.72 & 93.55 & 94.60 & \textbf{97.78} & \textbf{90.87} & 88.40 & 96.94 \\
LOGCAN++ & 78.58 & 85.34 & 86.62 &  87.51 &  93.76 &  77.20 &  79.78 &  93.13 \\
\hline
\textbf{Our SAIP-Net} & \textbf{88.88} & \textbf{93.79} & \textbf{93.97} & \textbf{94.84} & 97.66 & 90.85 & \textbf{89.49} & \textbf{96.99} \\
\bottomrule
\end{tabular}}

\end{table}

Extensive experimental evaluations, including comprehensive qualitative and quantitative analyses conducted on diverse remote sensing datasets, validate the superiority of SAIP-Net over baseline methods. Our method outperforms baseline approaches with comparable model complexity, demonstrating superior efficiency and effectiveness. The results clearly demonstrate the profound benefits and potential of integrating spectral-adaptive information propagation strategies.



\subsection{Analysis of Failure Cases}

\begin{figure}[!t]
    \centering
    \includegraphics[width=\linewidth]{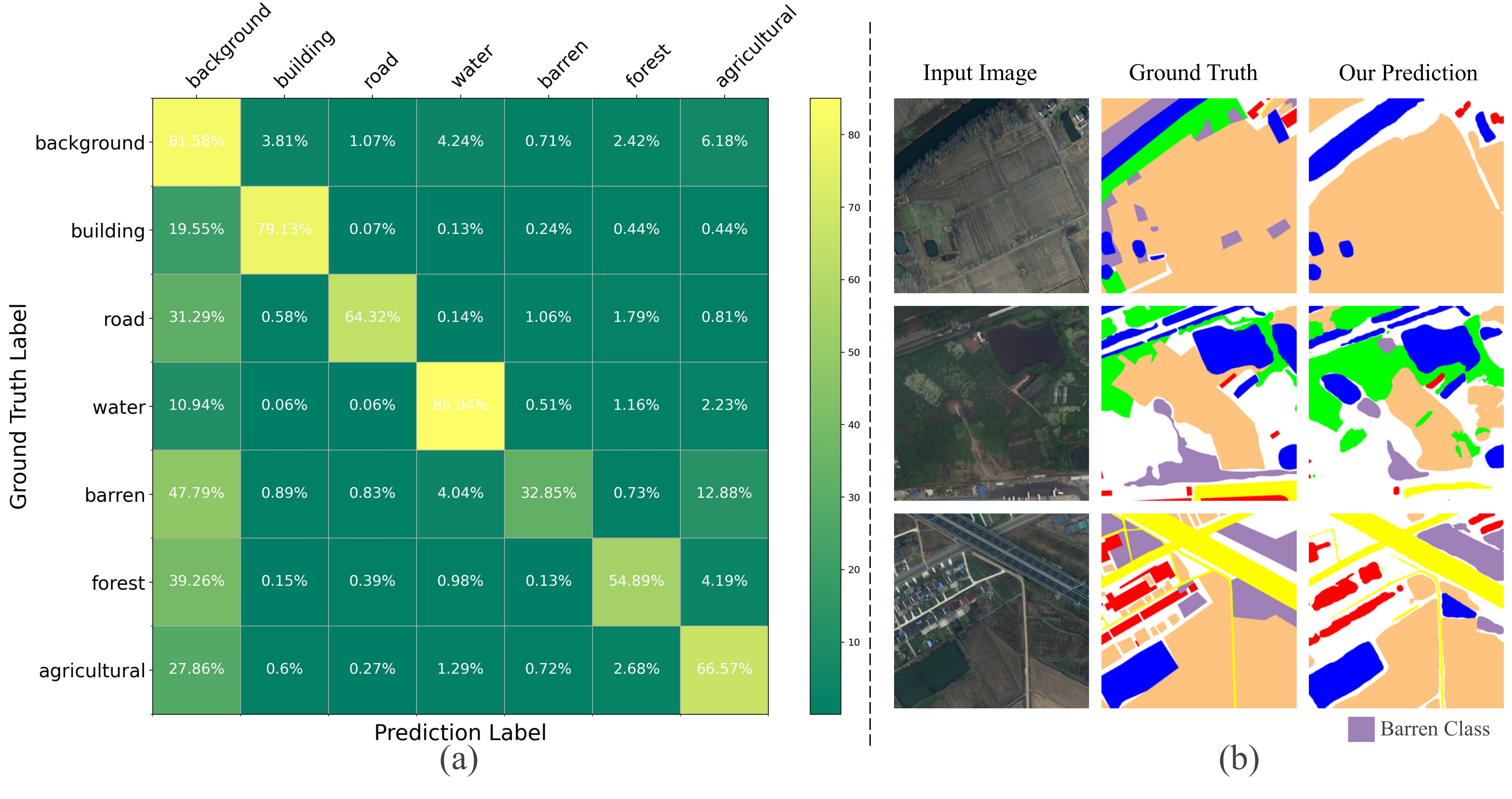}
    
    \caption{Analysis of failure cases on LoveDA Validation Set. (a). Class-wise \emph{normalized} confusion matrix, showing that the largest confusion involves the \textit{Barren} class, which is frequently predicted as \textit{agricultural} or \textit{background}. 
(b). The \textit{Barren} class (purple) illustrates typical errors such as boundary confusion with nearby man-made edges and “interior erosion’’ in low-texture regions.}
    \label{fig:analysis}
    
\end{figure}

Although SAIP-Net achieves consistent gains on most categories, we observe a noticeable drop for \textbf{Barren} on LoveDA dataset.

\paragraph{Data and appearance factors.}
As shown in Figure \ref{fig:analysis}, compared with other categories, Barren in LoveDA is intrinsically ambiguous under RGB due to low texture/contrast and appearance proximity to dry agriculture or background compounded by boundary-level annotation uncertainty. These conditions amplify class ambiguity and increase annotation noise near transitions. Together with the relatively smaller frequency of Barren pixels, the class becomes under-represented and more susceptible to bias during optimization.

\paragraph{Frequency bias induced by our design.}
SAIP-Net explicitly enhances high-frequency structures via LhpfStem and HP-guided paths in SAFF. While this helps boundaries, Barren regions are often low-texture, low-contrast. The HP-centric pathway may down-weight interior evidence in such regions, causing “interior erosion”. In addition, the LP branch in SAFF, normalized by softmax, favors smooth responses but is guided by features dominated by man-made edges in mixed scenes; this can pull Barren toward adjacent classes during the second-stage offset-based alignment. Finally, CDC enlarges the receptive field with dilations; when texture is scarce, multi-scale context may inadvertently absorb surrounding semantics, increasing contextual leakage into Barren.

Our design prioritizes boundary accuracy and intra-class consistency at controlled complexity, thus we deliberately avoid class-specific heavy treatments (e.g., large reweighting or post-processing), which leads to a moderate drop on this particular class. We also note a plausible side effect of our spectral priors: high-pass–oriented pathways can slightly erode interiors of textureless regions, and large receptive fields may absorb surrounding context. We view handling such low-texture, look-alike regions as a complementary future direction (e.g., class-aware calibration and mid-band constraints or multimodal cues), aiming to improve Barren without compromising efficiency.

\subsection{Ablation Study}

\begin{figure}[!t]
    \centering
    \includegraphics[width=\linewidth]{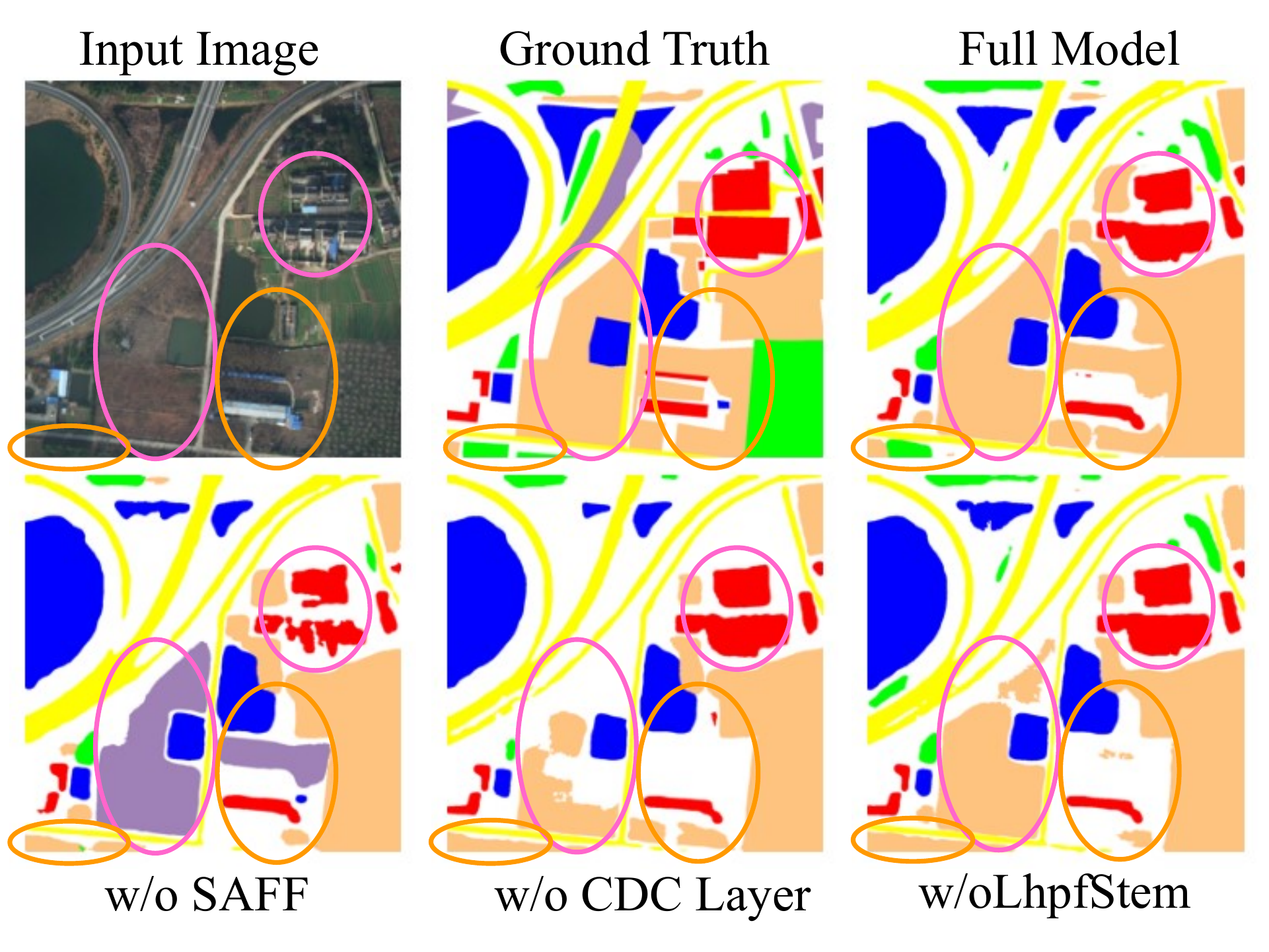}

    \caption{Qualitative comparison of ablation variants. The regions enclosed by pink ellipses indicate areas with improved intra-class consistency, and the areas marked by orange ellipses highlight enhanced boundary accuracy.}
    \label{fig:ablation_vis}

\end{figure}

To thoroughly investigate the effectiveness of each critical component in our proposed SAIP-Net, we conduct a detailed ablation study on Potsdam dataset and LoveDA validation set. 
For consistency, we use the same training configurations and evaluation metrics described in Section \ref{sec:impl_detail} and Section \ref{sec:eval_metrics}. Each ablation configuration is trained and tested independently. 

As shown in Table~\ref{tab:ablation}, progressively adding these modules into baselines improves segmentation performance across all metrics. Qualitative comparison of ablation is shown in Figure~\ref{fig:ablation_vis}. 
We can see that the combination of the SAFF module, CDC layers, and LhpfStem collaboratively enhances boundary quality, class consistency, multi-scale feature fusion, and edge information representation.

\begin{table}[t]
\centering
\caption{Ablation Study Results on the Potsdam dataset and LoveDA validation set. We use DeepLabV3+ as the baseline model, progressively replacing its components with our proposed modules. The symbol ``\checkmark'' indicates the inclusion of the corresponding component, while ``--'' denotes its absence. T.E indicates transformer encoder.}
\begin{tabular}{cccc|c|c}
\toprule
\multicolumn{4}{c|}{\textbf{Components}} & \multicolumn{2}{c}{mIoU$\uparrow$} \\
\midrule
T.E. & SAFF & CDC & LhpfStem & \textbf{LoveDA }&  \textbf{Potsdam}  \\
\midrule
-- & -- & -- & --  & 50.15  & 81.69  \\
\checkmark & -- & -- & -- & 52.85 &  87.76  \\
\checkmark & \checkmark  & -- & -- & 53.16  & 88.07 \\
\checkmark & \checkmark & \checkmark & -- & 54.07   & 88.34   \\
\checkmark & \checkmark & \checkmark & \checkmark & \textbf{54.46}  & \textbf{88.88}  \\
\bottomrule
\end{tabular}
\label{tab:ablation}
\vspace{1em}
\end{table}

We also ablate the proposed SAFF module to isolate its contribution to robustness. Specifically, we compare the full SAIP-Net against (i) an ablated variant \emph{w/o SAFF} that removes the module while keeping all other components, training protocol, and hyperparameters unchanged, and (ii) the baseline \emph{AerialFormer}. To probe robustness under common corruptions, we add synthetic Gaussian noise ($\mathrm{prob}{=}0.5$, $\sigma{=}10$), salt-and-pepper noise ($\mathrm{prob}{=}0.5$, ratio$=0.01$), and speckle noise ($\mathrm{prob}{=}0.5$, $\sigma{=}0.1$) at validation time. We report the \emph{drop} in mIoU relative to the clean setting (lower is better).

As summarized in Table~\ref{tab:noise_robustness}, SAIP-Net consistently exhibits the smallest performance degradation across all noise types. Relative to the baseline, SAIP-Net reduces the mIoU drop by $1.87$ (Gaussian), $1.50$ (Speckle), and $2.32$ (S\&P). Compared to the \emph{w/o SAFF} ablation, SAIP-Net further narrows the drop by $0.28$, $0.33$, and $0.74$ mIoU under Gaussian, Speckle, and S\&P noise, respectively. These gains align with SAFF’s design goal of suppressing disruptive high-frequency components while preserving task-relevant structures.

\begin{table}[h]
  \centering
  \caption{Performance drop in mIoU ($\downarrow$) under synthetic corruptions at validation time.
Values denote the decrease relative to the clean setting (lower is better). We compare SAIP\mbox{-}Net (Ours), its ablation w/o SAFF, and the AerialFormer baseline. Bold numbers indicate the best performance
in each row.}
  \label{tab:noise_robustness}
  \begin{tabular}{c|c|c|c}
  \hline
  \textbf{Noise} & \textbf{Ours} & \textbf{w/o SAFF} & \textbf{Baseline} \\
  \hline
  Gaussian & \textbf{0.56} & 0.84 & 2.43 \\
  Speckle  & \textbf{0.11} & 0.44 & 1.61 \\
  S\&P     & \textbf{2.47} & 3.21 & 4.79 \\
  \hline
  \end{tabular}
\end{table}


Additionally, we conduct ablation experiments on the hyperparameter $D$ of the CDC Layer using the LoveDA validation set. As shown in Table~\ref{tab:cdc_ablation}, the experimental results indicate that \(D=3\) achieves the best performance. Therefore, we adopt \(D=3\) in our final model.

\begin{table}[htbp]
  \centering
  \caption{Ablation Study on the Effect of \(D\) in CDC Layer on Model Performance. We conducted experiments on the LoveDA validation set and found that setting \(D=3\) yields the best performance.}
  \label{tab:cdc_ablation}
  \begin{tabular}{c|c|c|c|c}
    \hline
    The Value of D & $D=2$ & $D=3$ & $D=4$ & $D=5$ \\
    \midrule
    mIoU$\uparrow$ & 54.04 & \textbf{54.46} & 54.14 & 53.58  \\
    \bottomrule
  \end{tabular}
\end{table}

Our summarized ablation results clearly show the incremental performance improvement as each component is introduced. Furthermore, qualitative visualizations of segmentation maps illustrate specific improvements in boundary sharpness and intra-class consistency brought about by the combination of proposed modules. Through these experiments, we reinforce the rationale behind the design of SAIP-Net.

\section{Conclusion}

In this paper, we introduced SAIP-Net: Enhancing Remote Sensing Image Segmentation via \textbf{S}pectral \textbf{A}daptive \textbf{I}nformation \textbf{P}ropagation. By adaptively propagating spectral information and enlarging the receptive fields, our approach effectively eliminates disruptive high-frequency information within intra-class regions while preserving essential frequency-domain features along class boundaries. Experimental results confirm that our design significantly improves intra-class consistency and enhances boundary accuracy, thereby substantially improving the segmentation of complex structures in remote sensing images. A remaining limitation is reduced performance on inherently ambiguous, low-texture categories under RGB-only inputs (e.g., Barren in LoveDA dataset). As future work, we will explore methods to better disambiguate such regions without increasing complexity.




\end{document}